\documentclass[a4paper,twoside]{article}
\usepackage{hyperref}

\usepackage{amsmath,amssymb,amsfonts}
\usepackage{algorithmic}
\usepackage{graphicx}
\usepackage{textcomp}
\usepackage{xcolor}
\usepackage{multicol, multirow}
\usepackage{stfloats}
\usepackage{float}
\usepackage{appendix}
\usepackage{caption}
\usepackage{array}
\usepackage{colortbl}
\usepackage{tikz}
\usepackage{pgf}
\usepackage{pgfplots}
\usepackage{listings}
\usepackage{xcolor}
\usepackage{epsfig}
\usepackage{subcaption}
\usepackage{calc}
\usepackage{amssymb}
\usepackage{amstext}
\usepackage{amsmath}
\usepackage{amsthm}
\usepackage{multicol}
\usepackage{pslatex}
\usepackage{algorithm2e}
\usepackage{array}
\usepackage{changepage}
\usepackage[bottom]{footmisc}
\usepackage[backend=biber, style=apa, maxcitenames=2]{biblatex}
\usepackage{SCITEPRESS}     

\usepackage{hyperref}

\usepackage{amsmath,amssymb,amsfonts}
\usepackage{algorithmic}
\usepackage{graphicx}
\usepackage{textcomp}
\usepackage{xcolor}
\usepackage{multicol, multirow}
\usepackage{stfloats}
\usepackage{float}
\usepackage{appendix}
\usepackage{caption}
\usepackage{array}
\usepackage{colortbl}
\usepackage{tikz}
\usepackage{etoolbox}
\usepackage{pgf}
\usepackage{pgfplots}
\usepackage{listings}
\usepackage{xcolor}
\definecolor{codegray}{rgb}{0,0,0}
\definecolor{backcolour}{rgb}{0.88,0.92,1}

\lstdefinestyle{mystyle}{
    backgroundcolor=\color{backcolour},   
    numberstyle=\small\bfseries\color{codegray},
    basicstyle=\ttfamily\footnotesize,   
    breaklines=true,                 
    captionpos=b,              
    numbers=left,                  
    numbersep=3pt,
    tabsize=2,
    xleftmargin=0.25cm
}

\definecolor{high}{HTML}{1D6Df7}  
\definecolor{low}{HTML}{fc9432}   
\definecolor{neutral}{HTML}{FFFFFF} 

\newcommand*{\opacityorange}{50}
\newcommand*{\opacityblue}{49}
\newcommand*{\minval}{-0.8}
\newcommand*{\maxval}{0.8}

\newcommand{\gradient}[1]{
    \ifdimcomp{#1pt}{=}{0.0pt}{\cellcolor{neutral!\opacity} #1}{
        \ifdimcomp{#1pt}{>}{0.0pt}{
            \pgfmathparse{int(round(100*(#1/(\maxval))+(100)))}
            \xdef\tempa{\pgfmathresult}
            \cellcolor{high!\tempa!neutral!\opacityblue} #1
        }{
            \pgfmathparse{int(round(100*(#1/(\minval))+100))}
            \xdef\tempa{\pgfmathresult}
            \cellcolor{low!\tempa!neutral!\opacityorange} #1
        }
    }
}

\begin{document}

\title{Comparative Experimentation of Accuracy Metrics in Automated Medical Reporting: The Case of Otitis Consultations}

\author{\authorname{Wouter Faber$^*$\sup{1}\orcidAuthor{0009-0001-8412-9009}, Renske Eline Bootsma$^*$ \sup{1}\orcidAuthor{0009-0008-0667-2517}, Tom Huibers\sup{2}, Sandra van Dulmen\sup{3}\orcidAuthor{0000-0002-1651-7544} \\ and Sjaak Brinkkemper\sup{1}\orcidAuthor{0000-0002-2977-8911}}
\affiliation{\sup{1}Department of Information and Computing Sciences, Utrecht University, Utrecht, The Netherlands}
\affiliation{\sup{2}Verticai, Utrecht, The Netherlands}
\affiliation{\sup{3}NIVEL (Netherlands institute for health services research), Utrecht, The Netherlands}
\email{\{w.m.faber, r.e.bootsma, s.brinkkemper\}@uu.nl, tomhuibers@verticai.nl, s.vandulmen@nivel.nl}
}


\abstract{Generative Artificial Intelligence (AI) can be used to automatically generate medical reports based on transcripts of medical consultations. The aim is to reduce the administrative burden that healthcare professionals face. The accuracy of the generated reports needs to be established to ensure their correctness and usefulness. 
There are several metrics for measuring the accuracy of AI generated reports, but little work has been done towards the application of these metrics in medical reporting. 
A comparative experimentation of 10 accuracy metrics has been performed on AI generated medical reports against their corresponding General Practitioner's (GP) medical reports concerning Otitis consultations.
The number of missing, incorrect, and additional statements of the generated reports have been correlated with the metric scores. In addition, we introduce and define a Composite Accuracy Score which produces a single score for comparing the metrics within the field of automated medical reporting. 
Findings show that based on the correlation study and the Composite Accuracy Score, the ROUGE-L and Word Mover's Distance metrics are the preferred metrics, which is not in line with previous work. 
These findings help determine the accuracy of an AI generated medical report, which aids the development of systems that generate medical reports for GPs to reduce the administrative burden.    
}

\onecolumn \maketitle \normalsize \setcounter{footnote}{0} \vfill

\def\thefootnote{*}\footnotetext{These authors contributed equally to this work}

\section{\uppercase{Introduction}}
\label{sec:introduction}

\begin{quote}
    "Administrative burden is real, widespread and has serious consequences \parencite{heuer2022more}."
\end{quote}
\vspace{0.3cm}
Although the Electronic Health Record (EHR) has its benefits, the consequences regarding time and effort are increasingly noticed by medical personnel \parencite{olivares2019essential, 10.1093/jamia/ocaa325}. In addition to less direct patient care \parencite{lavander2016working}, documentation sometimes shifts to after hours: studies of \cite{anderson2020ehr} and \cite{saag2019pajama} show that physicians spend hours on documentation at home. Working after hours, a poor work/life balance, stress, and using Health Information Technology such as EHRs are associated with less work-life satisfaction and the risk of professional burnout \parencite{SHANAFELT2016836, robertson2017electronic, 10.1093/jamia/ocy145, hauer2018physician}.

The notation used for documentation of General Practitioner (GP) consultations is the Subjective, Objective, Assessment and Plan (SOAP) notation, which has been widely used for clinical notation, and dates back to 1968 \parencite{weed1968, HEUN1998235, sapkota2022community}. Based on a consultation, the SOAP note has to be written by the clinician, to update the EHR. To reduce the administrative burden, generative Artificial Intelligence (AI) can be used to summarize transcripts of medical consultations into automated reports, which is also the purpose of the Care2Report program to which this study belongs \parencite{molenaar2020, maas2020, kwint2023, wegstapel2023}. Care2Report (C2R) aims at automated medical reporting based on multimodal recording of a consultation and the generation and uploading of the report in the electronic medical record system \parencite{brinkkemper2022}. However, for these reports to be useful, the accuracy of the generated report has to be determined. Several metrics exist to compare the accuracy of generated text \parencite{Moramarco2022}, but the application in the Dutch medical domain has not been researched. 

\begin{table*}[b!h!]
\caption{Overview of existing accuracy metrics for Natural Language Generation.}

\begin{tabular}{|p{0.11\textwidth}|p{0.14\textwidth}|p{0.32\textwidth}|p{0.32\textwidth}|}
\hline
\textbf{Category} & \textbf{Metrics} & \textbf{Property} & \textbf{Common use$^{\mathrm{a}}$} \\
\hline
\multirow{4}{*}{Edit distance} & Levenshtein & Cosine similarity & MT, IC, SR, SUM, DG \& RG \\
 & WER & \% of insert, delete, and replace & SR \\
 & MER & Proportion word matches errors & SR \\
 & WIL & Proportion of word information lost & SR \\
\hline
\multirow{7}{*}{Embedding} & ROUGE-WE & ROUGE + word embeddings & SUM \\
 & Skipthoughts & Vector based similarity & MT \\
 & VectorExtrema & Vector based similarity & MT \\
 & GreedyMatching & Cosine similarity of embeddings & RG \\
 & USE & Sentence level embeddings & MT \\
 & WMD & EMD$^{\mathrm{b}}$ on words & IC \& SUM \\
 & BertScore & Similarity with context embeddings & DG \\
 & MoverScore & Context embeddings + EMD$^{\mathrm{b}}$ & DG \\
\hline
\multirow{7}{*}{Text overlap} & Precision & \% relevant of all text & MT, IC, SR, SUM, DG, QG, \& RG \\
 & Recall & \% relevant of all relevant & MT, IC, SR, SUM, DG, QG, \& RG \\
 & F-Score & Precision and recall & MT, IC, SR, SUM, DG, QG, \& RG \\
 & BLEU & $n$-gram precision & MT, IC, DG, QG, \& RG \\
 & ROUGE-n & $n$-gram recall & SUM \& DG \\
 & ROUGE-L & Longest common subsequence  & SUM \& DG \\
 & METEOR & $n$-gram with synonym match & MT, IC, \& DG \\
 & CHRF & $n$-gram F-score & MT \\
\hline
\multicolumn{4}{l}{$^{\mathrm{a}}$Abbreviations for the subfield, as introduced by \parencite{Celikyilmaz2020}. \textbf{MT}: Machine} \\ 
\multicolumn{4}{l}{Translation, \textbf{IC}: Image Captioning, \textbf{SR}: Speech Recognition, \textbf{SUM}: Summarization, \textbf{DG}: Document or} \\
\multicolumn{4}{l}{Story Generation, Visual-Story Generation, \textbf{QG}: Question Generation, \textbf{RG}: Dialog Response Generation.} \\
\multicolumn{4}{l}{$^{\mathrm{b}}$EMD = Earth Mover's Distance.}
\end{tabular}
\label{tab:overviewmetrics}

\end{table*}

Therefore, this paper proposes research towards metrics in the Dutch medical domain, resulting in the following research question: \newline

\textbf{RQ}
\textit{What is the preferred metric for measuring the difference between an automatically generated medical report and a general practitioner's report?} \newline

This study contributes to the field of AI generated medical reports by providing a case-level look and adds to the larger field of Natural Language Generation (NLG). Furthermore, this work has societal relevance by providing the preferred accuracy measure for AI-generated Dutch medical reports. Being able to identify the accuracy of a medical report, research towards the generation of the reports can be extended, to ensure a high accuracy. Namely, since reports play a crucial role in patient care, diagnosis, and treatment decisions it is vital that generated reports are correct and complete. Reports with high accuracy would prevent the medical staff from spending a lot of time writing the report themselves or correcting the generated reports, which reduces the administrative burden.

First, a literature review will be presented in \autoref{sec:relatedwork}. The method and findings will be discussed in \autoref{sec:method} and  \autoref{sec:findings} respectively, after which these will be discussed in \autoref{sec:discussion}. Conclusions will be drawn and directions for future work will be given in \autoref{sec:conclusion}.

\section{\uppercase{Related Work}}
\label{sec:relatedwork}
Different accuracy metrics for NLG exist, which can be compared in various ways.

\subsection{Accuracy Metrics} \label{sec:existingmetrics}
Over the years, different evaluation metrics for measuring the accuracy of NLG systems have been developed, such as BLEU \parencite{papineni2002}, ROUGE \parencite{lin2004}, and METEOR \parencite{banerjee2005}. All these metrics compare a generated text with a reference text. In recent years, a number of studies have provided an overview of these metrics and divided them into different categories \parencite{Sai2020, Celikyilmaz2020, Fabbri2020, Moramarco2022}. Our study adopts most of the metrics and categories of \cite{Moramarco2022}, which was inspired by the categories stated by \cite{Sai2020} and \cite{Celikyilmaz2020}. Metrics that are not specifically developed for summarization are also included, to ensure that the study does not become too narrow. \cite{Moramarco2022} also introduce a new metric with its own category, the Stanza+Snomed metric, which is not included in this current study since there is no other known work or use of this metric. The remaining three groups of metrics are: 

\begin{itemize}
    \item \textbf{Edit distance metrics} count how many characters or words are required to convert the output of the system into the reference text. They include Levenshtein \parencite{levenshtein1966}, Word Error Rate (WER) \parencite{su1992}, Match Error Rate (MER) \parencite{morris2004}, and Word Information Lost (WIL) \parencite{morris2004}. 

    \item \textbf{Embedding metrics} use encode units of text and pre-trained models to compute cosine similarity to find a similarity between the units. For this, they use word-level, byte-level, and sentence-level embeddings. The metrics include: ROUGE-WE \parencite{ng2015}, Skipthoughts \parencite{kiros2015}, VectorExtrema \parencite{forgues2014}, GreedyMatching \parencite{sharma2017}, Universal Sentence Encoder (USE) \parencite{cer2018}, Word Mover's Distance (WMD) \parencite{kusner2015}, BertScore \parencite{zhang2019bert}, and MoverScore \parencite{zhao2019}.

    \item \textbf{Text overlap metrics} rely on string matching, and measure the amount of characters, words, or $n$-grams that match between the generated text en the reference. These metrics include BLEU \parencite{papineni2002}, ROUGE \parencite{lin2004}, METEOR \parencite{banerjee2005}, and Character $n$-gram F-score (CHRF) \parencite{popovic2015}. The F-measure, based on precision and recall \parencite{maroengsit2019}, also falls under this category.
 
\end{itemize}

\autoref{tab:overviewmetrics} provides an overview of all the metrics, along with their category, property, and common use. This table extends data from \cite{Moramarco2022} and \cite{Celikyilmaz2020}. Information that was not available in these studies was derived from the original papers of the metrics.

\subsection{Comparison of Metrics}

In order to perform an evaluation of the accuracy metrics, a reference accuracy score is necessary to compare the calculated scores. There are different ways to determine these reference scores, which could involve a human evaluation. One could simply ask human evaluators to compare the generated text with a reference text and rate the general factual accuracy on a scale of 1 to 5 \parencite{Goodrich2019}. However, this is a very broad measure that is heavily influenced by subjectivity. Alternatively, other studies used different dimensions to compare generated texts, such as \textit{Adequacy}, \textit{Coherence}, \textit{Fluency}, \textit{Consistency}, and \textit{Relevance} \parencite{Fabbri2020, kryscinski2019, Turian2003}. \cite{Moramarco2022}, use \textit{Omissions}, \textit{Incorrect statements}, and \textit{Post-edit times} to evaluate automatically generated medical reports. 

The ratings of the human evaluators can be compared with the results of the metrics, using correlation measures such as the Spearman, Pearson, or Kendall's $\tau$ coefficient \parencite{Goodrich2019, Moramarco2022, Fabbri2020}.

\subsection{SOEP Reporting for GPs} \label{RW: domain}

In the Netherlands, the SOEP convention is used by GPs for medical reporting, which is the Dutch alternative to SOAP \parencite{van1996probleemlijst}. Subjective (S) represents the state, medical history and symptoms of the patient. Objective (O) contains measurable data obtained from past records or results of the examination and Evaluation (E) (or Assessment (A)) offers the opportunity to note the assessment of the health problem and diagnosis. Finally, Plan (P) contains the consultant's plan for the patient. Close attention should be paid to the division between the symptoms and signs (subjective descriptions and objective findings) since this is a common pitfall while writing SOEP notes \parencite{podder2022, seo2016}.

\section{\uppercase{Research Method}}
\label{sec:method}

\begin{figure}[h]
    \centering
    \includegraphics[width = 0.38\textwidth]{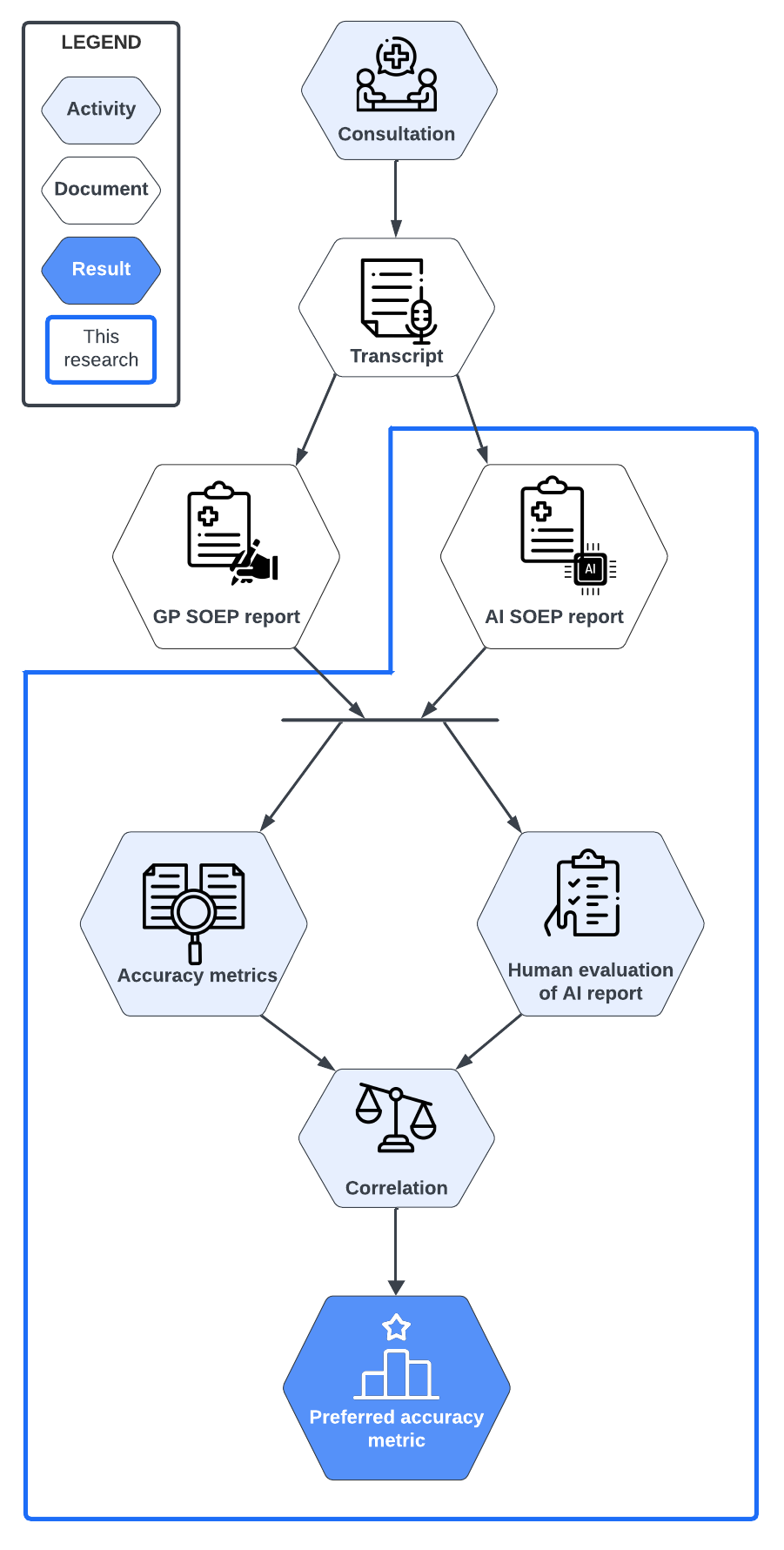}
    \captionsetup{font=footnotesize,justification=centering,labelsep=period}
    \caption{Research Method Diagram, showing the method along with its input and intended output.}
    \label{fig:RD}
\end{figure}

An overview of the research method of our study is shown in \autoref{fig:RD}, which will be explained in the subsections. The blue outline shows our research focus. 

\subsection{Materials} 
The C2R program provides data from seven transcripts of medical consultations between GPs and their patients concerning ear infections, namely \textit{Otitis Externa} ($n$ = 4) and \textit{Otitis Media Acuta} ($n$ = 3) \parencite{maas2020}. These transcripts are derived from video recordings, for which both the patients and GPs provided informed consent. The recordings were made as part of a study by Nivel (Netherlands institute for health services research) and Radboudumc to improve GP communication \parencite{houwen2017, meijers2019}. 

Based on the transcripts, GPs wrote a SOEP report (referred to as GP report), which is considered the ground truth for this study. These GPs did not perform the consultation but wrote the report solely based on the transcripts. Furthermore, software of the C2R program which runs on GPT 4.0 was used. The temperature was set to 0 to limit the diversity of the generated text. Based on the formulated prompt and transcript, the GPT generates a SOEP report (referred to as AI report) \parencite{maas2020}.

\subsection{Pre-study}
Upon first inspection of the GP reports, it was noticed that abbreviations such as "pcm" meaning "paracetamol" are frequently used. To gain more insights into the experience and preferences of medical staff regarding the formulation of SOEP reports, a pre-study was conducted among Dutch medical staff ($n$ = 5; 1 physiotherapist, 1 paediatrician, 1 junior doctor / medical director, 1 nurse and 1 nursing student). The participants were asked about their experience with (SOEP) medical reporting, important factors of SOEP, the use of abbreviations, and general feedback on the Care2Report program. All participants indicated to have knowledge of SOEP reporting, and have experience with writing medical reports, using SOEP or similar methods. Distinguishing between Subjective and Objective information was indicated to be a common mistake, which is in line with research \parencite{podder2022, seo2016}. In addition, the notation of the Evaluation is important since this is "the essence of the consult", but is sometimes not filled in completely. Regarding abbreviations, the participants were divided. Some of them indicated that they preferred using abbreviations, to enable faster reading, but discouraged the use of difficult abbreviations. The other participants indicated always favouring written terms since this improves readability. All participants favoured using written terms when multiple staff members (from different backgrounds) were involved, for example when it came to patient transfer, with the exception of general abbreviations.

In general, the medical staff agreed that an AI report would be "a great solution" that "saves time, which enables more consultation time". In addition, two of the medical staff indicated writing reports after the consult due to time limits, which can cause a loss of information. This insight is in line with previous research \parencite{olivares2019essential, 10.1093/jamia/ocaa325, lavander2016working, anderson2020ehr, saag2019pajama}.

\subsection{Prompt for Report Generation}
The GPT software does not have the knowledge or capability to use medical abbreviations like GPs use in their SOEP report. This can result in the metrics falsely identifying differences between written terms and their abbreviations. That, in combination with the fact that using written terms was preferred by half of the medical staff of the pre-study and since the reports will be read by staff members from different disciplines, led to the decision to change abbreviations in the GP's report to the full expression.

The GPT was given a Dutch prompt, of which the translated text is given in \autoref{prompt}. The formulation of the prompt was based on existing research within the C2R program (line 1, 2, 3, 4, 5, 7, 9), and has been adapted to incorporate the input of the medical staff (line 3, 4, 5, 6, 8) and literature (line 3, 4, 5). Mainly, the division between symptoms and signs (Subjective and Objective) and the definition of the Evaluation category have been added.

\begin{lstlisting}[caption = Prompt used as input for the GPT., label = prompt]
Write a medical s.o.e.p report based on a conversation between a gp and a patient and use short and concise sentences. 
Report in the categories of subjective, objective, evaluation, and plan. 
Make sure that for subjective the description of the complaints of the patient is noted.
Also, possible pain medication which is used by the patient and the information that emerges from the anamnesis may be noted here.
At objective, the observation of the symptoms by the gp and the results of the physical examination must be noted.
The evaluation contains the judgement of the examination and the diagnosis.
The treatment plan must be clear from the plan. 
Make sure that the medical terms, such as the name of the medication are noted.
The content of the report must be derived from the given transcript.
\end{lstlisting}

\subsection{Metric Selection and Execution} 
For this study, a spread of metrics between categories (see \autoref{sec:existingmetrics}) was chosen. More popular or common metrics were preferred due to frequent application and public availability. The following 10 metrics are part of the selection: \textit{Levenshtein}, \textit{WER}, \textit{ROUGE-1}, \textit{ROUGE-2}, \textit{ROUGE-L}, \textit{BLEU}, \textit{F-Measure}, \textit{METEOR}, \textit{BertScore}, \textit{WMD}.

Each accuracy metric is applied to the AI report with the GP report as reference. Five of the metrics could be run via an online application and the F-Measure was calculated using an R function. In addition, the embedding metrics, \textit{BertScore} and \textit{WMD}, required running Python code. For these metrics, Dutch embeddings were used. BertScore supports more than 100 languages via multilingual BERT, including Dutch, and for WMD the "dutch-word-embeddings" Word2Vec model was used \parencite{Nieuwenhuijse_2018}. METEOR uses $n$-gram comparison with synonym match. At the time of writing, there is no alternative for Dutch texts. Therefore, METEOR will mostly rely on $n$-gram comparison.

\subsection{Human evaluation} \label{sec: method human eval}
Concurrently, the AI reports are compared with the GP reports by the first authors, i.e., the human evaluation. This is inspired by the work of \cite{Moramarco2022}. This method of evaluation is adopted because it is a domain-specific method that includes the accuracy of the report and provides insight into the amount of work needed by the GP as well. For each AI report, seven aspects will be counted, as can be seen in \autoref{tab:humanaspects}. 

\begin{table}[h]
\caption{Human evaluation aspects along with their descriptions and abbreviations. \centering}
\label{tab:humanaspects} \centering
\begin{adjustwidth}{-0.55cm}{}
\begin{tabular}{|>{\footnotesize}l|>{\footnotesize}l|>{\tiny}l|}
  \hline
  \textbf{Aspect}  & \textbf{Description} & {\footnotesize \textbf{Abr.}}\\
  \hline
  Missing &  Missing in AI report & MIS\\
  Incorrect  &  Incorrect in AI report & INC\\
  Added On-topic  & Not in GP report, on-topic & $ADD_{ON}$ \\
  Added Off-topic  & Not in GP report, off-topic & $ADD_{OFF}$\\
  Post-edit time & Time ($s$) to correct AI report & PET\\
  Nr. of characters & Nr. of characters in AI report & NRC\\
  Word length & Avg. word length in AI report & WLE \\
  \hline
\end{tabular}
\end{adjustwidth}
\end{table}

Firstly, the number of \textit{Missing statements} and \textit{Incorrect statements}, which include wrongly stated information. Next, the \textit{Additional statements}, which are divided into \textit{On-topic} and \textit{Off-topic}. \textit{Added On-topic} statements contain information that is not present in the GP report but relates to the content, e.g "There is no pus visible, but there is blood leaking from the ear". \textit{Added Off-topic} statements contain information that is not present in the GP report and does not relate to the content, e.g. "The patient called in sick for work". In addition, the \textit{Number of characters} and the number of words will be counted, to calculate the \textit{Word length}. An independent samples $t$-test will be performed on the \textit{Number of characters} and the \textit{Word length} between the AI report and the GP report to gain more insight into a potential difference in report length. Lastly, the \textit{Post-edit time} describes the time it takes to correct the AI report, i.e., adding \textit{Missing}, changing \textit{Incorrect} and removing \textit{Additional} statements. This is interesting to consider since the goal of AI reporting is to reduce the time spent on reporting by GPs.

After performing the human evaluation, Pearson correlation coefficients are calculated between the aspects of the human evaluation and every metric, excluding the \textit{Word length}, and \textit{Number of characters}. In theory, the stronger the negative correlation, the more effective the metric is in the domain of medical reporting. Namely, an AI report ideally has low missing statements, low incorrect and low additional statements. 

To compare the metrics, a single \textit{Composite Accuracy Score} \textit{(CAS)} is calculated for each metric. For this, the correlations per \textit{Missing}, \textit{Incorrect} and \textit{Added} statements with the metric are normalised on a scale from 0 to 1, where 0 is the lowest (negative) correlation and 1 is the highest. Based on the normalised correlations with the \textit{Missing (MIS)}, \textit{Incorrect (INC)}, \textit{Added On-topic ($ADD_{ON}$)}, and \textit{Added Off-topic $(ADD_{OFF})$} statements, the \textit{Composite Accuracy Score} \textit{(CAS)} is calculated using Formula \ref{casformula}.

\begin{equation}
\label{casformula}
CAS \: = \frac{MIS + INC + ADD_{OFF} + 0.5 \times ADD_{ON}}{3.5}
\end{equation}

Every score has a weight of 1.0, except the \textit{Added On-topic} statements, which have been attributed a weight of 0.5 since their presence in the AI report is deemed less severe than the other aspects. The \textit{Post-edit time} is not part of the \textit{Composite Accuracy Score} because it is dependent on the other aspects of the human evaluation. 

With respect to editing, a metric can be considered preferred if it has a low \textit{Composite Accuracy Score} as well as a strong negative correlation with the \textit{Post-edit time}. If a metric fulfils these requirements, it is an adequate tool to measure the accuracy of the report itself as well as the administrative burden.  

\begin{table*}[b!h!]
\caption{Human evaluation aspects per AI report. The averages of the aspects are rounded to whole numbers.}
\begin{center}
\begin{tabular}{|l|c|c|c|c|c|c|c||c|}
\hline
\textbf{Human Evaluation Aspects}  & \textbf{R1} & \textbf{R2} & \textbf{R3} & \textbf{R4} & \textbf{R5} & \textbf{R6} & \textbf{R7} & \textbf{Average}\\
\hline
Missing statements &12 &5 &8 &7 &9 &8 &7 & 8 \\
Incorrect statements &2 &1 &1 &2 &1 &5 &1 & 2 \\
Added statements - On-topic &6 &9 &5 &7 &8 &3 &7 & 6 \\
Added statements - Off-topic &5 &5 &0 &2 &1 &5 &0 & 3 \\
Post-edit time (s) &378 &196 &170 &213 &193 &186 &169 & 215 \\
\hline
\end{tabular}
\label{tab:human evaluation}
\end{center}
\end{table*}

\section{\uppercase{Findings}}
\label{sec:findings}

\begin{table*}[b!h!]
\caption{Pearson correlation between human evaluation aspects and metrics, along with the \textit{Composite Accuracy Score}.}
\begin{center}
\begin{tabular}{|p{2cm}|l|l|l|l||c||c|}
\hline
 \multirow{2}{*}{\textbf{Metric}} & \multirow{2}{*}{\textbf{Miss.}} & \multirow{2}{*}{\textbf{Incorr.}} & \multicolumn{2}{c||}{\textbf{Additional}}  & \multirow{2}{*}{\textbf{CAS}}  & \multirow{2}{*}{\textbf{PET}} \\
\cline{4-5}
 & & & \textbf{On-topic} & \textbf{Off-topic} & &  \\
\hline
\textit{Levenshtein} & \gradient{ 0.122} & \gradient{-0.178} & \gradient{-0.011}  & \gradient{-0.798}  & \textbf{0.229} & \gradient{  -0.320} \\
\textit{WER} & \gradient{ 0.673} & \gradient{-0.042} & \gradient{-0.409} & \gradient{-0.315}  & 0.434 & \gradient{   0.385}\\
\hline
\textit{BertScore} & \gradient{-0.272} & \gradient{ 0.126} & \gradient{ 0.319} & \gradient{ 0.759} &  0.618 &  \gradient{           0.329} \\
\textit{WMD} & \gradient{-0.564} & \gradient{-0.168} & \gradient{ 0.381} & \gradient{-0.289}  & \textbf{0.241} & \textbf{\gradient{-0.591      }}\\
\hline
\textit{ROUGE-1} & \gradient{-0.063} & \gradient{ 0.123} & \gradient{-0.394} & \gradient{-0.634}  & 0.284 & \textbf{\gradient{-0.483 }}\\
\textit{ROUGE-2} & \gradient{-0.153} & \gradient{ 0.131} & \gradient{-0.021} & \gradient{-0.259} &  0.401 & \gradient{ -0.201} \\
\textit{ROUGE-L} & \gradient{-0.233} & \gradient{-0.109} & \gradient{-0.056} & \gradient{-0.597} & \textbf{0.209} &\textbf{\gradient{-0.461}} \\
\textit{BLEU} & \gradient{ 0.109} & \gradient{-0.013} & \gradient{ 0.002} & \gradient{-0.462} &  0.364 & \gradient{  -0.258} \\
\textit{F-Measure} & \gradient{ 0.698} & \gradient{ 0.119} & \gradient{-0.434} & \gradient{-0.339} &  0.501 & \gradient{  0.333}\\
\textit{METEOR} & \gradient{ 0.315} & \gradient{ 0.467} & \gradient{-0.220} & \gradient{ 0.045} &  0.677 & \gradient{   0.103}\\
\hline
\multicolumn{7}{l}{The negative correlations are indicated by different intensities of orange, and the positive}\\
\multicolumn{7}{l}{correlations are indicated by different intensities of blue. The three lowest Composite} \\
\multicolumn{7}{l}{Accuracy Scores and PET correlations are in \textbf{bold}.} 

\end{tabular}
\label{tab:maintable}
\end{center}
\end{table*}

Performing the method resulted in measured human evaluation aspects, correlations and the \textit{Composite Accuracy Scores}.

\subsection{Human Evaluation}
As mentioned in \autoref{sec: method human eval}, the seven human evaluation aspects were counted for each AI report, of which the first five can be seen in \autoref{tab:human evaluation}. Often, the ear in question was \textit{Missing} in the AI report. \textit{Incorrect} statements were statements which were wrongly stated or wrongly defined as being said by the GP. Of the statements that were not in the GP report (\textit{Added}), the distinction between \textit{On-topic} and \textit{Off-topic} was less direct. Mainly, \textit{On-topic} statements contained additional information regarding the medical history, complaints or treatment. Statements regarding other topics then discussed in the GP report and explanations to the patient were classified as \textit{Off-topic} because these would not be of any relevance to the SOEP report, written by the GP.

The \textit{Number of characters} and the number of words were used to calculate the \textit{Word length} for both the GP report and AI report. The results of the independent samples $t$-test show that the AI reports are significantly longer in terms of characters $(1199.29 \pm 197)$ than the GP report $(410.71 \pm 94.32), t(12) = -9.520, p <0.001$. The words used in the AI report $(6.04 \pm 0.33)$ are significantly shorter than in the GP report $(7.62 \pm 0.29), t(12) = 9.555, p < 0.001.$

\subsection{Correlation between Metrics and Human Evaluation Aspects}
Using the metric scores and the human evaluation aspects, the mutual correlation has been calculated. In contrast to the other metrics, for the edit distance metrics (Levenshtein and WER) a low score equals a good accuracy. To enable easier comparison between the metrics, the correlations of the edit distance metrics have been inverted by multiplying with -1. All correlations between the metrics and human evaluation aspects are shown in \autoref{tab:maintable}. Ideally, these correlations are strongly negative since the medical report should be concise and contain all, and only, relevant information. The three strongest negative correlations with the \textit{Post-edit time} and the three lowest \textit{Composite Accuracy Scores} have been indicated in bold in \autoref{tab:maintable}

\section{\uppercase{Discussion}}
\label{sec:discussion}
Based on the findings in \autoref{sec:findings}, notable observations were found.

\paragraph{Human Evaluation} The results of the human evaluation (\autoref{tab:human evaluation}) show that all AI reports contain on average 9 \textit{Added} statements, compared to the GP report. Consequently, the AI reports are longer than the GP reports. Most \textit{Added} statements are \textit{On-topic}. Despite the additional information and length of the AI report, each report misses on average 8 statements.

\paragraph{Added statements} The first noticeable correlation in \autoref{tab:maintable} appears between more than half of the metrics correlating moderately (-0.3 $< r <$ -0.5) or strongly ($r <$ -0.5) negatively with \textit{Added Off-topic} statements. Interestingly, only three metrics have a moderate negative correlation with \textit{Added On-topic} statements. This can be explained by the \textit{On-topic} statements adding extra, relevant, information to the content of the SOEP report. Even though these statements are added, the metrics could define this as relevant information thus not having a strong negative correlation.

\paragraph{Comparison with \cite{Moramarco2022}} Except for the WMD metric, none of the metrics strongly correlate negatively with \textit{Missing} statements and \textit{Post-edit time}, which is not in line with the findings of \cite{Moramarco2022}. In their findings, METEOR and BLEU scored good on detecting \textit{Missing} statements and Levenshtein and METEOR rank highly on the \textit{Post-edit time}. Additionally, none of the metrics moderately or strongly correlate negatively with \textit{Incorrect} statements, which is also not in line with the findings of \cite{Moramarco2022}, where ROUGE scored good on identifying these statements.

\paragraph{Opposite of preferred correlations} The WER and F-Measure strongly correlate ($r >$ 0.5) \textit{Missing} statements with better accuracy, and BertScore correlates a high number of \textit{Off-topic} statements with better accuracy. These results indicate exactly the opposite of what is preferred and therefore seem to be less suitable for the evaluation of automatically generated reports. 

\paragraph{Post-edit time} Six metrics have a negative correlation with the \textit{Post-edit time}. WMD, ROUGE-1, and ROUGE-L have the strongest negative correlation, meaning that they are the preferred metrics concerning the correlation with \textit{Post-edit time}.

\paragraph{Composite Accuracy Score} When looking at the \textit{CAS}, The WER, BertScore, F-Measure and METEOR metrics score high ($>$ 0.5), indicating that these metrics are not suitable for the current application. The high \textit{CAS} of the BertScore is remarkable since this metric performed as one of the best in the study of \cite{Moramarco2022}. The high \textit{CAS} of METEOR could be explained due to the fact that the used transcripts are Dutch, which is an unsupported language by the metric. Therefore, it cannot use synonym matching, which is the added benefit of the METEOR metric compared to other text overlap metrics. There is no consensus within the categories of edit distance, embedded and text overlap metrics. Consequently, no conclusions can be drawn regarding preferred performing categories.  

\paragraph{Preferred metrics} Based on the \textit{CAS} and the \textit{Post-edit time} correlations, ROUGE-L and WMD are the preferred metrics, since they are in the top 3 for both. The WMD scores slightly worse in terms of \textit{CAS}, which can be explained by the fact that it has a positive correlation (0.381) with the \textit{Added On-topic} statements, whereas ROUGE-L has only negative correlations with the human evaluation metrics. However, WMD scores better than ROUGE-L when looking at the \textit{Post-edit time}. \cite{Moramarco2022} found that Levenshtein, BertScore, and METEOR are the most suitable metrics, which does not correspond with the findings of our work.

\section{\uppercase{Conclusion}}
\label{sec:conclusion}
AI generated medical reports could provide support for medical staff. These reports should be as accurate as possible, to limit the time needed by the medical staff making corrections. To determine the accuracy of a text, metrics can be used. This research investigated the performance of 10 accuracy metrics by calculating the correlation between the metric score and the following human evaluation aspects: \textit{Missing} statements, \textit{Incorrect} statements, \textit{Additional} statements and \textit{Post-edit time}.

For each metric, the \textit{Composite Accuracy Score} has been calculated, indicating its performance. Based on the \textit{CAS} and the correlation with the \textit{Post-edit time}, the \textit{ROUGE-L} and \textit{Word Mover's Distance (WMD)} metrics are preferred to use in the context of medical reporting, answering the research question: 

\textit{What is the preferred metric for measuring the difference between an automatically generated medical report and a general practitioner's report?}

Based on the results, we see that there is a diversity among the applications of the different metrics. Both strong positive and negative correlations with the human evaluation aspects are found, which can be explained by the different methods used by the metrics. The preferred method depends heavily on the context of use and which aspect is deemed more important. Therefore, no unambiguous answer can be given. However, we created the \textit{CAS} score based on our context of use, identifying the preferred metrics in the context of medical reporting. 

\subsection{Limitations}
The outcome of this study is not in line with previous research, which could be due to the limitations. There are three main limitations to this study. 

Firstly, the data set used for running the accuracy metrics consists of just seven AI reports. Additionally, the transcripts used are all from GP consultations on \textit{Otitis Externa} and \textit{Otitis Media Acuta}, making the data limited in its medical diversity. These factors make it difficult to draw general conclusions on accuracy metrics that work for all AI generated medical reports.

Adding to that, the GP reports were written solely based on the transcripts, and not by the GP who performed the consultation, which is not standard practice.

Lastly, the human evaluation was performed by researchers who have no prior experience in writing medical reports. Even though medical staff was consulted in this study, it would be preferred if the human evaluation was done by people with medical expertise. That way, those with a deeper comprehension of what should be included in a report could handle the more challenging cases of evaluating the generated statements' relevance.

\subsection{Future Work}
The main limitations should be addressed in future work. Mainly, the study should be repeated with more medical reporting, on other pathologies. Besides, it would improve the quality of the study if the human evaluation were executed by healthcare professionals. Furthermore, the current AI reports result in low accuracy scores for each metric. Therefore, it would be beneficial if further research was done into optimising the prompt formulation, resulting in more accurate AI reports. Additionally, the human evaluation of this study does not take wrongly classified statements to the SOEP categories into account, which could be adopted in future work. Finally, the use of abbreviations in generated reports could be further explored, since this was taken out of the equation for this study.  

\section*{\uppercase{Acknowledgements}}
Our thanks go to the medical staff who helped us with the pre-study. In addition, the icons of Flaticon.com enabled us to create \autoref{fig:RD}. Finally, many thanks go to Bakkenist for the support of this research project.

\renewcommand\refname{REFERENCES}
\AtNextBibliography{\small}

{
\printbibliography}

\end{document}